%% file: main.tex
\documentclass[10pt,twocolumn,letterpaper]{article}

\usepackage{iccv}              

\definecolor{iccvblue}{rgb}{0.21,0.49,0.74}
\usepackage[pagebackref,breaklinks,colorlinks,allcolors=iccvblue]{hyperref}
\usepackage{multirow}
\usepackage{makecell}
\usepackage{diagbox}
\usepackage{algorithm} 
\usepackage{algorithmicx}
\usepackage{algpseudocode}
\usepackage{graphicx}
\usepackage{amsmath}
\usepackage{mathtools}  

\def\paperID{1656} 
\def\confName{ICCV}
\def\confYear{2025}

\title{EvolvingGrasp: Evolutionary Grasp Generation via
\\ Efficient Preference Alignment}

\author{
Yufei Zhu$^{1,*}$,
Yiming Zhong$^{1,*}$,
Zemin Yang$^{1}$,
Peishan Cong$^{1}$,\\
Jingyi Yu$^{1}$, 
Xinge Zhu$^{2,\dagger}$,
Yuexin Ma$^{1,\dagger}$
\\ 
$^{1}$ ShanghaiTech University $^{2}$ The Chinese University of Hong Kong \\
\textcolor{red}{\url{https://evolvinggrasp.github.io/}}
}

\begin{document}
\makeatletter
\let\@oldmaketitle\@maketitle

\renewcommand{\@maketitle}{
   \@oldmaketitle
 \begin{center}
\includegraphics[width=1\linewidth]{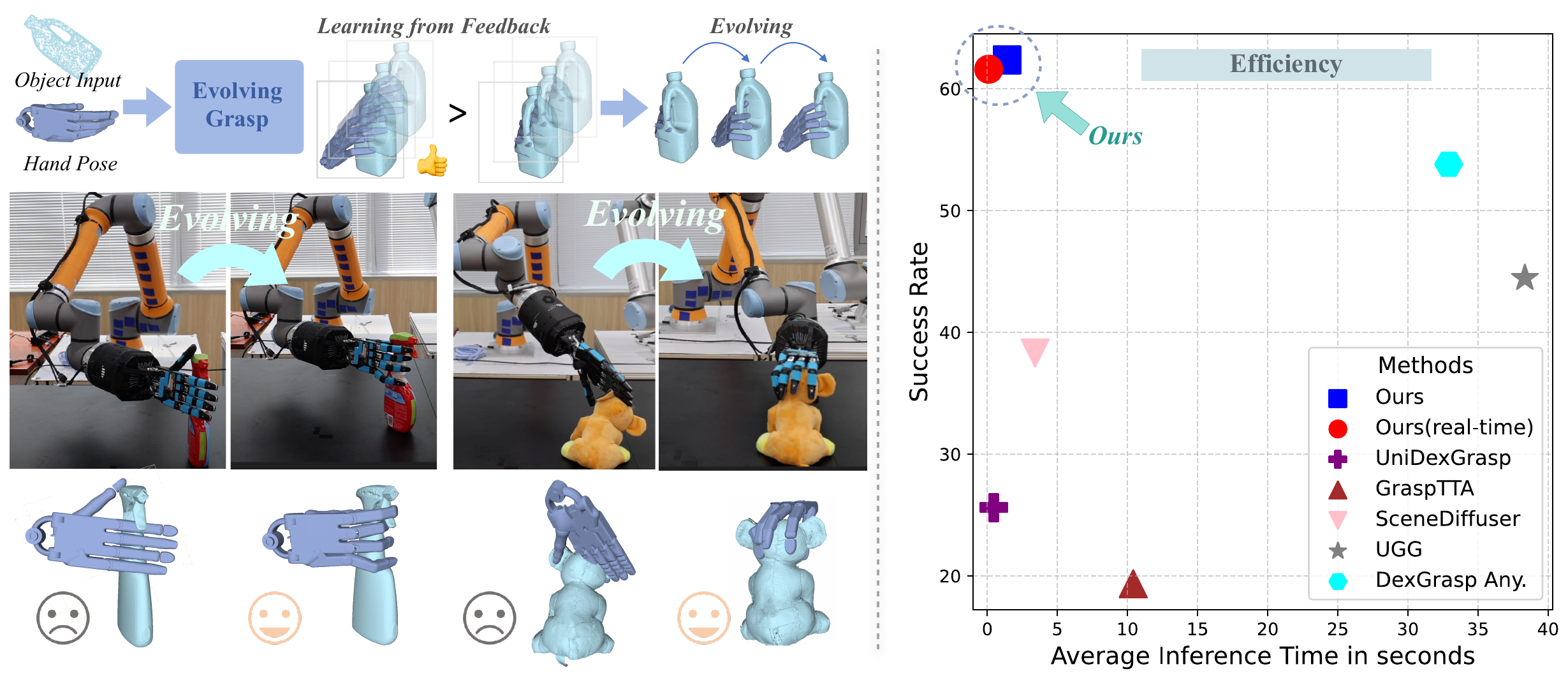}
 \end{center}
 \vspace{-3ex}
  \refstepcounter{figure}\normalfont Figure~\thefigure. 
  The left part illustrates \textbf{EvolvingGrasp}, an approach akin to evolution, where it enables the model to learn from experience and iteratively refine its grasping strategy. The right part demonstrates its efficiency and effectiveness
  \label{fig:teaser}
  \newline
  }

\maketitle
\footnotetext[1]{$^{*}$ Equal contribution.}
\footnotetext[2]{$\dagger$ Corresponding author. This work was supported by NSFC (No.62206173), Shanghai Frontiers Science Center of Human-centered Artificial Intelligence (ShangHAI), MoE Key Laboratory of Intelligent Perception and Human-Machine Collaboration (KLIP-HuMaCo).}

\input{sec/0_abstract}

\input{sec/1_intro}

\input{sec/2_related_work}
\input{sec/3_method}
\input{sec/4_exp}

\input{sec/5_conclusion}

{
    \small
    \bibliographystyle{ieeenat_fullname}
    \bibliography{main}
}

\appendix
\input{sec/X_suppl}

\end{document}

%% file: sec/0_abstract.tex
\begin{abstract}

Dexterous robotic hands often struggle to generalize effectively in complex environments due to the limitations of models trained on low-diversity data. However, the real world presents an inherently unbounded range of scenarios, making it impractical to account for every possible variation. A natural solution is to enable robots learning from experience in complex environments—an approach akin to evolution, where systems improve through continuous feedback, learning from both failures and successes, and iterating toward optimal performance. 
Motivated by this, we propose EvolvingGrasp, an evolutionary grasp generation method that continuously enhances grasping performance through efficient preference alignment.   
Specifically, we introduce Handpose-wise Preference Optimization (HPO), which allows the model to continuously align with preferences from both positive and negative feedback while progressively refining its grasping strategies. To further enhance efficiency and reliability during online adjustments, we incorporate a Physics-aware Consistency Model within HPO, which accelerates inference, reduces the number of timesteps needed for preference fine-tuning, and ensures physical plausibility throughout the process.
Extensive experiments across four benchmark datasets demonstrate state-of-the-art performance of our method in grasp success rate and sampling efficiency. Our results validate that EvolvingGrasp enables evolutionary grasp generation, ensuring robust, physically feasible, and preference-aligned grasping in both simulation and real scenarios.

\end{abstract}

%% file: sec/1_intro.tex
\section{Introduction}


Dexterous robotic grasping has made significant strides in embodied manipulation, enabling more adaptive and precise interactions compared to traditional grippers. Existing grasping methods~\cite{ponce1993characterizing, rosales2012synthesis, dai2018synthesis, ferrari1992planning, zhang2024dexgrasp, liu2021synthesizing,liu2019generating, romero2022embodied, li2024multi, liu2020deep, huang2024fungrasp, zhang2024graspxl, zhang2024artigrasp, weng2024dexdiffuser,shao2024bimanual} generally fall into two categories: optimization-based approaches~\cite{ponce1993characterizing, rosales2012synthesis, ferrari1992planning, dai2018synthesis}, which refine hand poses to achieve force-closure states, learning-based approaches~\cite{lundell2021ddgc, wei2022dvgg, wei2024grasp,xu2024dexterous}, which directly map object features to grasp poses through regression, probabilistic modeling, and generative-based methods~\cite{scenediffuser, grasptta, lu2023ugg, lundell2021multi,weng2024dexdiffuser}, which utilize diffusion model to estimate the distribution of hand poses. Recent advances, such as DexGrasp Anything~\cite{graspanyting}, have further introduced physics-based constraints to improve grasp feasibility. However, a fundamental limitation persists—limited generalization. These methods, trained on limited datasets, struggle to adapt to complex environments. 
This challenge is exacerbated by an inherent property of the real world: \textbf{its unbounded diversity}. The vast range of object shapes, materials, and environmental conditions makes it impractical to predefine an exhaustive set of grasping strategies. Without the ability to adapt in deployment, grasping models remain constrained, failing to handle variations beyond their training distribution. To overcome this, a natural approach is to enable evolutionary grasp generation, where the system \textbf{learns from experience} (\ie, both failures and successes), refining its grasping strategy through iterative improvements based on real-world interactions. This process not only enhances generalization but also allows for preference alignment, ensuring that grasping behaviors adapt to task-specific requirements and user-defined preferences. However, achieving efficient evolutionary refinement is non-trivial, as many existing learning-based approaches rely on slow, compute-intensive updates, particularly in diffusion-based models that require numerous iterative steps and physics simulations.

To address these challenges, we propose EvolvingGrasp, an evolutionary grasp generation framework that efficiently refines grasp strategies through preference alignment while maintaining physical plausibility. At its core, we introduce \textbf{Handpose-wise Preference Optimization} (HPO), a novel method that reformulates preference alignment~\cite{wallace2024diffusion,zhou2025dreamdpo,zhang2024grape,tan2024sopo, chen2025fdpp} as a posterior probability optimization problem, encouraging generated grasps to converge toward preferred distributions while diverging from non-preferred ones. Notably, the proposed HPO is an extension of Direct Preference Optimization (DPO)~\cite{rafailov2023direct}, where it is also, to the best of our knowledge, the first to incorporate the DPO into the dexterous grasp. To further improve efficiency, we integrate HPO into a \textbf{Physics-Aware Consistency Model} (including two parts, \ie, \textbf{Physics-Aware Distillation} for training and \textbf{Physics-Aware Sampling} for inference), which pretrains a diffusion model and distills it into a lightweight, few-step sampling model. This enables both \textbf{rapid inference and efficient preference fine-tuning}, significantly reducing the number of required sampling steps and optimization iterations. Additionally, we introduce {three physics-aware constraints} to ensure the stability, realism, and feasibility of generated grasp poses—surface pulling force to maintain stable contact, external penetration repulsion force to prevent object penetration, and self-penetration repulsion force to avoid inter-finger collisions. 

Extensive experiments across four benchmark datasets demonstrate that EvolvingGrasp achieves state-of-the-art results, significantly improving grasp success rate, sampling efficiency, and physical plausibility, with \textbf{30x speedup} over existing methods, demonstrating robust generalization across simulated and real-world benchmarks. Our contributions can be summarized as follows:

\begin{itemize}
\item We introduce EvolvingGrasp, an efficient evolutionary grasp generation framework that enables iterative refinement, addressing the challenge of generalizing to diverse and unstructured real-world environments.
\item Efficient preference alignment is achieved through Handpose-wise Preference Optimization (HPO), which formulates grasp adaptation as a posterior probability optimization problem, enabling the model to iteratively converge toward preferred grasp distributions.
\item We propose a Physics-Aware Consistency Model (PCM) that accelerates preference alignment by reducing sampling steps while enforcing geometric consistency and physical feasibility through structured constraints.
\item Extensive experiments across four benchmark datasets demonstrate that our method achieves state-of-the-art grasp success rates, physical plausibility, and sampling efficiency. Furthermore, it enables real-time grasp generation with minimal computational overhead, achieving comparable performance to gradient-based methods.

\end{itemize}






\label{sec:intro}

%% file: sec/2_related_work.tex
\begin{figure*}[t]
	\centering
    \includegraphics[width=1.0\linewidth]{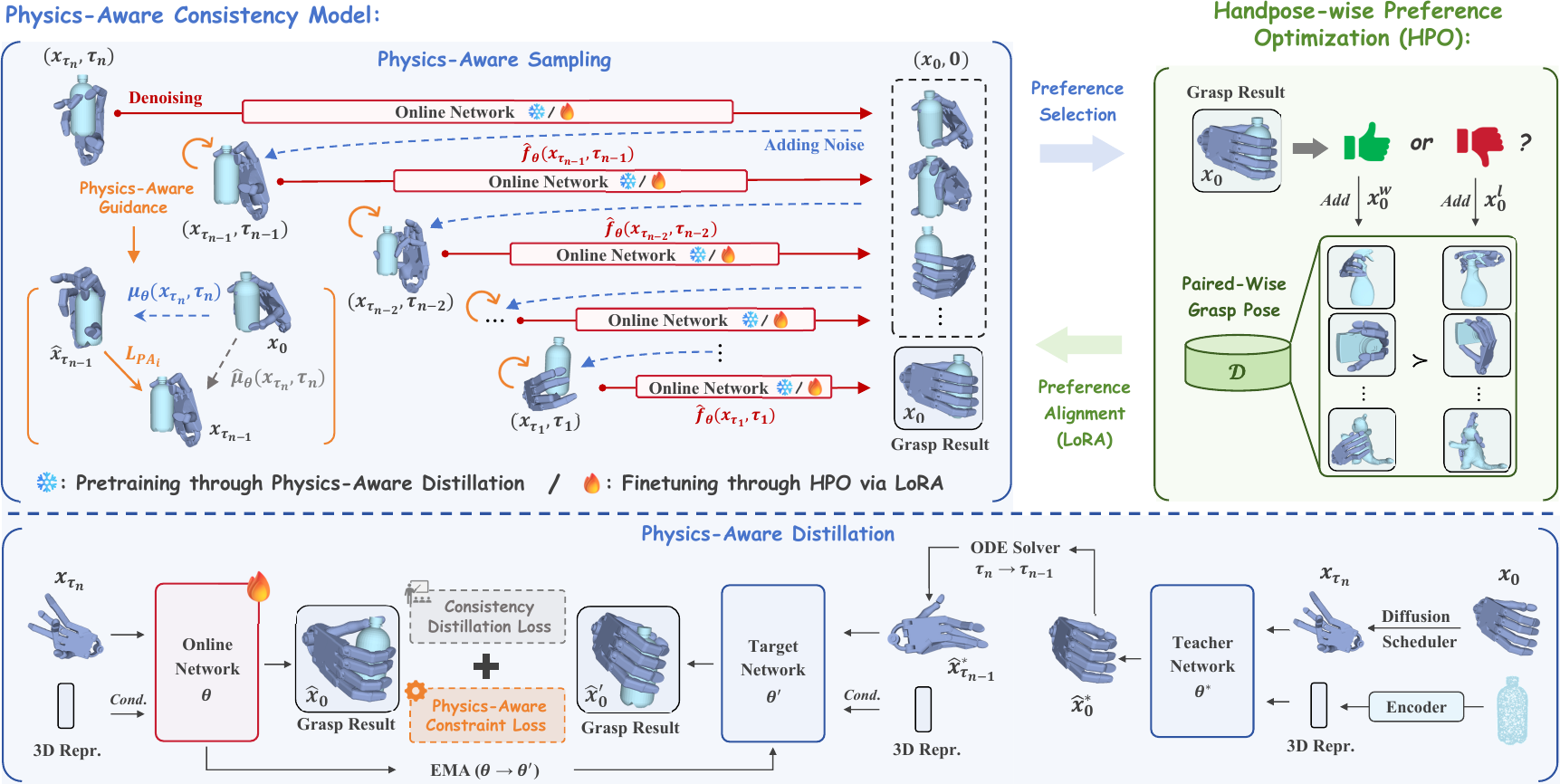}
	\centering
	\caption{Overview of EvolvingGrasp. The evolutionary process begins with the human preference guidance, where Handpose-wise Preference Optimization (HPO, highlighted in the green rectangle) is employed to facilitate preference alignment. These grasp poses are generated by the Physics-Aware Consistency Model (shown in blue rectangles), including Sampling and Distillation mechanism, to ensure the sampling efficiency and the physical plausibility. In this way, EvolvingGrasp, an efficient evolutionary grasp generation framework is proposed to enable the grasp model iteratively converge toward preferred distributions.}
    \label{pipeline}
\end{figure*}


\section{Related Work}

\subsection{Dexterous Grasping}

Dexterous grasp generation aims to produce diverse and high-quality grasping poses for robotic hands to interact with objects effectively. Recent works can be categorized into regression-based~\cite{liu2019generating, romero2022embodied, liu2020deep} and generation-based methods~\cite{scenediffuser, grasptta, lu2023ugg, lundell2021multi, lundell2021ddgc, wei2022dvgg, wei2024grasp}. Regression-based methods directly predict grasping parameters from the input object, but they often suffer from mode collapse issues that limits output diversity. Generation-based methods, though capable of producing varied solutions, face efficiency challenges. SceneDiffuser~\cite{scenediffuser}, UGG~\cite{lu2023ugg}, and DexGrasp Anything \cite{graspanyting} require multiple sampling steps to generate diverse grasping poses, and DexGrasp Anything~\cite{graspanyting} additionally incurs computational overhead by calculating physical constraint losses and performing gradient updates at each sampling iteration. 
However, existing methods cannot generate grasping poses aligned with human habits due to their lack of evolutionary adaptation and preference alignment. 
They also struggle to balance physical plausibility with computational efficiency.

\label{realted_Dexterous_Grasping}

\subsection{Accelerating Diffusion Models}

Accelerating dexterous grasping is vital for enhancing real-time adaptability. A series of diffusion-based acceleration methods have emerged, making real-time and efficient generation possible.
DDIM~\cite{song2020denoising} accelerates inference by reformulating the stochastic denoising process into a deterministic ODE solver. The DPM-Solver~\cite{lu2022dpm++,lu2022dpm} series further designs efficient high-order ODE solvers, achieving comparable quality with only $10\sim 20$ steps. However, their performance degrades significantly when sampling steps are reduced to $2\sim 4$. Consistency works~\cite{kim2023consistency, lu2024simplifying} support few-step sampling while preserving quality.  CTM~\cite{kim2023consistency} extends CM to reduce cumulative errors and discretization inaccuracies. sCMs~\cite{lu2024simplifying} unifies flow matching and diffusion frameworks to avoid discretization errors and hyperparameter tuning. Therefore,
we employ the consistency model to enable efficient evolutionary refinement, reducing inference timesteps while maintaining result quality. 



\subsection{Preference Finetuning for Diffusion Models}
Preference alignment play a crucial role inenabling generative models to learn from user feedback, optimize generation strategies, and progressively improve output quality.
Some researchers utilize it into diffusion models to better align with human preferences in image generation area. 
These methods can be categorized into two types, finetuning with reward model \cite{black2023training,  fan2023dpok, clark2023directly, deng2024prdp} and reward-free finetuning \cite{hiranaka2024human, yang2024using, wallace2024diffusion}.
The former includes DDPO~\cite{black2023training} and DPOK \cite{fan2023dpok} which treat the denoising process of diffusion models as an MDP and finetune using multiple reward models.
Finetuning without reward model includes D3PO~\cite{yang2024using} and Diffusion-DPO \cite{wallace2024diffusion}, which
extend the theoretical framework of DPO~\cite{rafailov2023direct} to multi-step MDPs. These approaches learn an optimal reward model and then use it to refine the sampling strategy, making them a relatively more cost-effective alternative.
However, RL-based fine-tuning for diffusion models still requires backpropagation at every sampled timestep, making it highly time-consuming. Applying this approach directly to grasping tasks is impractical for real-world applications, where efficient refinement is crucial.
Therefore, we propose a faster preference alignment fine-tuning approach that reduces backpropagation steps, improving grasping performance while aligning it with human preferences.

\label{sec:related}

%% file: sec/3_method.tex

\section{Methodology}

\subsection{Overview}

To achieve efficient evolutionary grasp generation that aligns with human preferences while maintaining both efficiency and physical plausibility, we propose EvolvingGrasp, as illustrated in Fig.~\ref{pipeline}. The evolutionary process begins with preference alignment, where we introduce Handpose-Wise Preference Optimization (HPO) (Section \ref{Self-Improving}). HPO formulates grasp adaptation as a posterior probability optimization problem, allowing the model to iteratively converge toward preferred grasp distributions. However, directly applying HPO suffers from slow sampling and inefficiencies in preference alignment, limiting its practicality for real-time grasp refinement. To address these challenges, we incorporate a Physics-Aware Consistency Model (PCM) (Section \ref{Physics_aware_Consistency_Model}), which leverages a consistency model framework to enhance efficiency by reducing the number of required sampling steps. While this improves inference speed, naive consistency-based sampling may still generate physically implausible grasp poses. To ensure geometric consistency and physical feasibility, PCM integrates a physics-aware distillation and sampling mechanism, which enforces structured physical constraints on the generated poses.

\subsection{Problem Formulation}
Given the object point cloud representation $O \in \mathbb{R}^{N\times 3}$, our goal is to generate dexterous grasp poses with high success rate and low penetration from the posterior distribution $P(x \mid O)$, where $x=\left \{ x_{i} \right \}_{i=1}^{n}$. Specifically, the pose parameters contain three parts, joint angles of the hand $\theta_h \in \mathbb{R}^{24}$, global translation $T_{global} \in \mathbb{R}^{3}$, and global rotation $R_{global} \in SO(3)$. 

Given ground truth samples from the data distribution $\pi(x_{0})$, the noise schedule weight $\alpha$, the goal of diffusion models is to fit the GT data distribution. The objective of training diffusion model is as follows:
\begin{equation}
\begin{aligned}
\mathbb{E}_{t \sim [0,T], x_{0}, \epsilon \sim N(0,I)}\left[\left\|\epsilon-\epsilon_{\theta}\left(x_{t}, t,O\right)\right\|^{2}\right]
\end{aligned}
\end{equation}
where $x_{t}=\sqrt{\bar{\alpha}_{t}} {x}_{0}+\sqrt{1-\bar{\alpha}_{t}}\epsilon $. Given the noise related parameter $\sigma_{t}$ and the mean from $x_{t}$ to $x_{t-1}$, the reverse process of diffusion model is as follows:
\begin{equation}
\begin{aligned}
p_{\theta}\left({x}_{t-1} \mid {x}_{t}\right)=\mathcal{N}\left({x}_{t-1} ; {\mu}_{\theta}\left({x}_{t}, t\right), \sigma_{t}^{2}{\rm I}\right)
\end{aligned}
\end{equation}

\label{Problem_Formulation}
\subsection{Handpose-wise Preference Optimization}
To achieve preference alignment in the evolutionary adaptation process, we introduce the HPO, an extension of direct preference optimization (DPO). In DPO, the fundamental assumption is that we have access to data generated by the model, with human annotators providing the corresponding preferences. Specifically, for a given object $O$, we observe pairs of grasp poses $x_{0}^{w}$ and $x_{0}^{l}$, where the set $x_{0}^{w}$ is preferred over $x_{0}^{l}$. These preferences can be modeled using the Bradley-Terry model, which expresses the probability of one sample being preferred over another as:
\begin{equation}
\begin{aligned}
p_{BT}\left(x_{0}^{w} \succ x_{0}^{l}\right)=\sigma\left(r\left(c, x_{0}^{w}\right)-r\left(c, x_{0}^{l}\right)\right)
\end{aligned}
\end{equation}

The training objective of DPO is to maximize the likelihood of the observed preferences, which can be formulated as a binary classification problem. Specifically, the whole model is finetuned to minimize the following loss function:
\begin{equation}
\begin{aligned}
L_{\mathrm{BT}}(\theta) 
= -\mathbb{E}_{(x_0^w, x_0^l) \sim \mathcal{D}} \log \sigma \Bigg( 
\quad  \mathop{\beta\mathbb{E}}_{\substack{
    x_{1:T}^w \sim \pi_\theta(x_{1:T}^w \mid x_0^w) \\
    x_{1:T}^l \sim \pi_\theta(x_{1:T}^l \mid x_0^l)
}} \\\left[ 
    \log \frac{\pi_\theta(x_{0:T}^w)}{\pi_{\mathrm{ref}}(x_{0:T}^w)} 
    - \log \frac{\pi_\theta(x_{0:T}^l)}{\pi_{\mathrm{ref}}(x_{0:T}^l)} 
\right] \Bigg)
\end{aligned}
\label{loss_BT}
\end{equation}
where $\mathcal{D}$ is the paired-wise grasp pose dataset. Following \cite{wallace2024diffusion}, we can get the upper bound of Eq. (\ref{loss_BT}) as:
\begin{equation}
\begin{aligned}
L_{BT}(\theta) \leq- \underset{\left({x}_{0}^{w}, {x}_{0}^{l}\right) \sim \mathcal{D}, n \sim \mathcal{U}(1, N), 
x_{{n-1}, {n}}^{w} \sim \pi_{\theta}\left(x_{{n-1}, {n}}^{w} \mid x_{0}^{w}\right)}{\beta\mathbb{E}} \\
\log \sigma\left( \log \frac{\pi_{\theta}\left({x}_{{n-1}}^{w} \mid {x}_{{n}}^{w}\right)}{\pi_{\mathrm{ref}}\left({x}_{{n-1}}^{w} \mid {x}_{{n}}^{w}\right)}- \log \frac{\pi_{\theta}\left({x}_{{n-1}}^{l} \mid {x}_{{n}}^{l}\right)}{\pi_{\mathrm{ref}}\left({x}_{{n-1}}^{l} \mid {x}_{{n}}^{l}\right)}\right)
\label{loss_upperbound}
\end{aligned}
\end{equation}
where $\pi_{\theta}\left({x}_{{n-1}} \mid {x}_{n}\right)$ is the probability of sampling ${x}_{{n-1}}$ from ${x}_{n}$, which can be computed by:
\begin{equation}
\begin{aligned}
\pi_{\theta}\left({x}_{{n-1}} \mid {x}_{{n}}\right) = 
\frac{1}{\sqrt{2 \pi} \sigma_{{n}}} \text{exp} (-\frac{({x}_{{n}}-\mu_{\theta }({x}_{{n}},{n}))^{2}}{2\sigma_{{n}}^{2}} )
\end{aligned}
\label{computing_probability}
\end{equation}

As far as we know, HPO is the first to integrate DPO into grasp pose generation, from which we extend DPO to a more flexible form. In HPO, there is no requirement to maintain an equal number of preferred and non-preferred grasp poses, which enables more adaptive preference learning. The objective of HPO is introduced as follows:
\begin{equation}
\begin{aligned}
\mathcal{L}_{HPO} = \underset{{x}_{0}^{i} \sim \mathcal{D}, n \sim \mathcal{U}(1, N), 
x_{{n-1}, {n}}^{i} \sim \pi_{\theta}\left(x_{{n-1}, {n}}^{i} \mid x_{0}^{i}\right)}{\beta\mathbb{E}} 
\log \sigma \\ \left(\sum_{i=1}^{N_{suc}} \log \frac{\pi_{\theta}\left({x}_{{n-1}}^{i} \mid {x}_{{n}}^{i}\right)}{\pi_{\mathrm{ref}}\left({x}_{{n-1}}^{i} \mid {x}_{{n}}^{i}\right)}- \sum_{j=1}^{N_{fail}}\log \frac{\pi_{\theta}\left({x}_{{n-1}}^{j} \mid {x}_{{n}}^{j}\right)}{\pi_{\mathrm{ref}}\left({x}_{{n-1}}^{j} \mid {x}_{{n}}^{j}\right)}\right)
\label{hpo_loss_upperbound}
\end{aligned}
\end{equation}
where $N_{suc}$ and $N_{fail}$ represent the number of preferred and non-preferred poses.
HPO optimizes grasp pose generation by quantifying the probabilistic divergence between successful grasps (preferred samples) and failed grasps (non-preferred samples), thereby driving the model towards human-preferred behaviors. This process inherently involves the dynamic adjustment and maximization of reward signals: preferred grasp poses are reinforced due to their higher probability, while non-preferred poses are suppressed, guiding the model to progressively discard ineffective strategies.

Preferred grasp selection can be conducted through either simulation-based evaluation or human-in-the-loop selection. In the simulation-based approach, poses that achieve success across all six directional evaluations are classified as preferred samples, while the remaining poses are treated as negative samples. Alternatively, in the human selection process, grasp poses that align with human intuition and habitual preferences are designated as preferred samples, while the others are considered non-preferred.
Finally, we finetune the whole model via LoRA~\cite{hu2022lora} using  Eq. (\ref{hpo_loss_upperbound}) to align the model with preferences.

\label{Self-Improving}

\subsection{Physics-Aware Consistency Model}
\label{Physics_aware_Consistency_Model}
\subsubsection{Consistency Model}
Directly applying HPO faces two major efficiency challenges: generating poses requires over hundreds of timesteps per inference, and preference fine-tuning demands a large number of backpropagation steps. To address these inefficiencies, we adopt the consistency model framework to accelerate both sampling process and preference alignment.
The core idea of the consistency model is to learn a mapping from any point along the ODE trajectory back to its starting point, which corresponds to the data distribution.
Given a Probability Flow ODE trajectory \cite{song2020score} $\left\{x_{\tau_{n}}\right\}_{\tau_{n} \in[0, T]}$, a consistency model $f_{\theta}$ is defined as:
\begin{equation}
\begin{aligned}
f_{\theta}:\left(x_{\tau_{n}}, \tau_{n}, O\right) \mapsto x_{0}
\end{aligned}
\end{equation}
where $x_{0}$ is the initial point of the trajectory, $O$ is the observation as condition. Due to space limitations, we omitted condition $O$ in other parts of the paper, except for the appendix. The self-consistency property ensures that:
\begin{equation}
\begin{aligned}
f_{\theta}\left(x_{\tau_{n}}, \tau_{n}\right)=f_{\theta^{\prime}}\left(x_{\tau_{n}^{\prime}}, \tau_{n}^{\prime}\right) \quad \forall \tau_{n}, \tau_{n}^{\prime} \in[0, T]
\end{aligned}
\end{equation}
The consistency model needs to meet a key boundary condition: when $t=0$, the model output should be the input itself, that is, $f_{\theta}(x_{0}, 0, O)=x_{0}$. This can be achieved as the following way:
\begin{equation}
\begin{aligned}
f_{\theta}(x_{\tau_{n}}, \tau_{n})=c_{\text {skip }}(\tau_{n}) x_{\tau_{n}}+c_{\text {out }}(\tau_{n}) F_{\theta}(x_{\tau_{n}}, \tau_{n})
\end{aligned}
\end{equation}
where $c_{skip}(t)$ and $c_{out}(t)$ are differentiable functions that satisfy $c_{skip}(\epsilon) = 1$ and $c_{out}(\epsilon) = 0$. $F_{\theta}(x, t)$ denotes a deep neural network that predicts $\hat{x}_{0}$.
\begin{equation}
\begin{aligned}
F_{\theta}(x_{\tau_{n}}, \tau_{n})=
\frac{1}{\sqrt{\bar{\alpha}_{\tau_{n}}}} \left(x_{\tau_{n}}-\sqrt{1-\bar{\alpha}_{\tau_{n}}} \epsilon_{\theta}\left(x_{\tau_{n}}, \tau_{n}\right)\right)
\end{aligned}
\end{equation}
Consistency distillation leverages pre-trained diffusion model to condense multi-step sampling into a more efficient few-step inference process. The process involves generating pairs of adjacent points on the PFODE trajectory using numerical ODE solvers and minimizing the difference between the model's outputs for these pairs. The loss function for consistency distillation is defined as:
\begin{equation}
\begin{aligned}
\mathcal{L}_{CD}=\mathbb{E}\left[d\left(f_{\theta}\left(x_{\tau_{n}}, \tau_{n}\right), f_{\theta^{\prime}}\left(\hat{x}_{\tau_{n-1}}^{\ast }, \tau_{n-1}\right)\right)\right]
\end{aligned}
\label{consistency_distillation_loss}
\end{equation}
where $O$ is omitted for simplicity, $\hat{x}_{\tau_{n-1}}^{\ast }$ is computed using a numerical ODE solver, $d(\cdot, \cdot)$ is a $L_2$ distance metric, $\theta$ and $\theta^{\prime}$ represent the online network and target network parameters. In addition, the parameters of $\theta^{\prime}$ are updated by the exponential moving average (EMA) of the parameters of $\theta$. $\hat{x}_{\tau_{n-1}}^{\ast}$ can be computed as:
\begin{equation}
\begin{aligned}
\hat{x}_{\tau_{n-1}}^{\ast} \longleftarrow \sqrt{\bar{\alpha}_{\tau_{n-1}}}F_{\theta}(x_{\tau_{n}}, \tau_{n}) + \sqrt{1-\bar{\alpha}_{\tau_{n-1}}} \epsilon 
\end{aligned}
\label{ddim_solver}
\end{equation}
During sampling, we utilize the consistency function to directly generate the final sample and the quality of the generated sample can be enhanced through an iterative process that alternates between denoising and injecting noise. Given the sequence of timesteps $S \in \left\{\tau_{i} \mid \tau_{0}=0, \tau_{N-1}=T, \tau_{i}<\tau_{i+1} \text { for } i=0,1, \ldots, N-1\right\}$, the adding-noise process can be formulated as:
\begin{equation}
\begin{aligned}
\hat{x}_{\tau_{n-1}}= \mu_{\theta}( {x}_{\tau_{n}},\tau_{n}) +\sigma_{\tau_{n}}{\epsilon}
\end{aligned}
\label{reverse_process}
\end{equation}
where $\mu_{\theta}=\sqrt{\bar{\alpha}_{\tau_{n-1}}}f_{\theta}(x_{\tau_{n}}, \tau_{n})$, $\sigma_{\tau_{n}}=\sqrt{1-\bar{\alpha}_{\tau_{n-1}}}$. Subsequently, we conduct the prediction of the final sample utilizing the trained consistency function once more.
\label{consistency_model}
\begin{figure*}[ht!]
	\centering
        \includegraphics[width=1.0\linewidth]{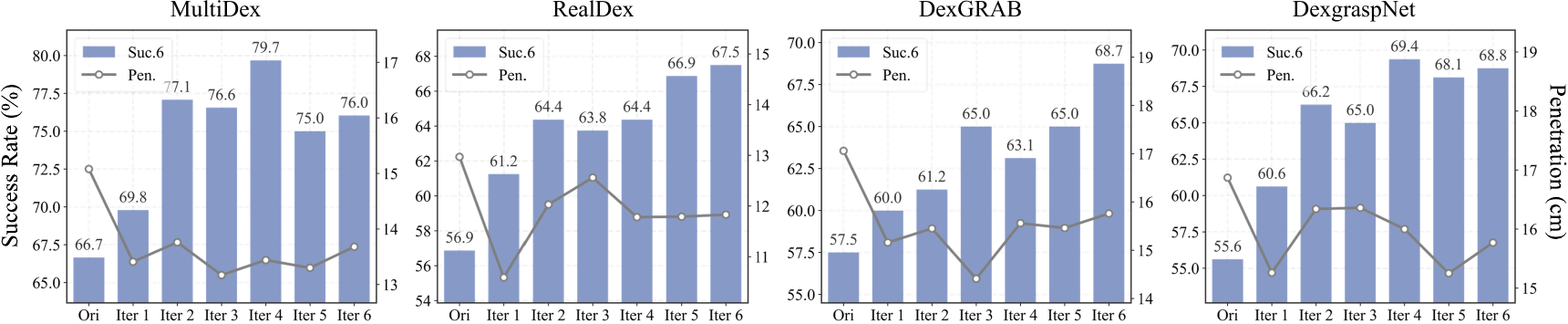}
	\centering
	\caption{Mean grasping performance in terms of success rate and penetration of randomly selected 6 objects with the finetuning epoch increasing during inference optimization.}
    \label{DPO_6_performance}
    \vspace{-3ex}
\end{figure*}
\subsubsection{Physics-Aware Distillation and Sampling}
Although adopting the consistency model improves sampling and preference fine-tuning efficiency, it still generates physically implausible poses. To address this, we introduce a Physics-Aware Distillation and Sampling paradigm that enforces physical constraints during the distillation process of  predicting $\hat{x}_{0}$ while ensuring that the sampling trajectories adhere to specific constraints. Following \cite{graspanyting}, we incorporate three physics-aware objectives, with the distillation objective of the consistency model formulated as follows:
\begin{equation}
\begin{aligned}
\mathcal{L}_{PAD}=\mathcal{L}_{CD}+
\sum_{i=1}^{m} \alpha_{i} L_{PA_{i}}&\left(F_{\theta}(x_{\tau_{n}}, \tau_{n}), \epsilon_{\theta}\right)
\end{aligned}
\label{physical_aware_disti_loss}
\end{equation}
where $L_{PA_{i}}\left(F_{\theta}(x_{t}, t), \epsilon_{\theta}\right)$ is the $i^{th}$ physical constraint loss and $m=3$ denotes three constraints, \ie, Surface Pulling Force, External-penetration Repulsion Force, Self-Penetration Repulsion Force respectively \cite{graspanyting}. 
These constraints guarantee grasping feasibility and maintain geometric accuracy in finger-to-object and inter-finger interactions.
$\alpha_{i}$ is the corresponding weight parameter. We first train a diffusion model for grasp pose generation as a teacher model. Then, we utilize $\mathcal{L}_{\mathrm{PAD}}$ to distill the teacher model into a student model.

During sampling, the consistency model estimates a clearer hand pose $x_{0}$ based on the current noisy hand pose $x_{\tau_{n}}$ and object information $O$ by consistency function $f_{\theta}(x_{\tau_{n}}, \tau_{n}, O)$. Subsequently, physical constraints are applied to steer the sampling process, making it closer to a physically feasible grasping pose. Following \cite{dhariwal2021diffusion, yang2024guidance,chung2022diffusion}, the gradient of the constraint loss is used to modify the mean from $x_{t}$ to $x_{t-1}$:
\begin{equation}
\begin{aligned}
\hat{\mu}_{\theta}(x_{\tau_{n}}, \tau_{n}) = \mu_{\theta}(x_{\tau_{n}}, \tau_{n}) \\
+ \sum_{i=1}^{m} \gamma_{i} \nabla_{x_{\tau_{n}}} L_{PA_{i}}&\left( F_{\theta}(x_{\tau_{n}}, \tau_{n}), \epsilon_{\theta} \right)
\end{aligned}
\label{compute_mean}
\end{equation}

where $\gamma_{i}$ is the weight parameter corresponding to each physical constraint. Moreover, we can derive a new mapping from noise to data as follows:
\begin{equation}
\begin{aligned}
\hat{f}_{\theta}(x_{\tau_{n}}, \tau_{n}) = f_{\theta}(x_{\tau_{n}}, \tau_{n})
+ \frac{1}{\sqrt{\bar{\alpha}_{\tau_{n-1}}}}\sum_{i=1}^{m} \gamma_{i} \nabla_{x_{\tau_{n}}} L_{PA_{i}}
\end{aligned}
\end{equation}
By employing the physics-aware consistency model, we derive a novel preference alignment objective based on the new sampling path. The detailed derivation is provided in Appendix B. As a result, our method can efficiently generate higher-quality poses during the evolutionary process.

\label{sec:methods}

%% file: sec/4_exp.tex
\begin{figure*}[ht!]
	\centering
        \includegraphics[width=1.0\linewidth]{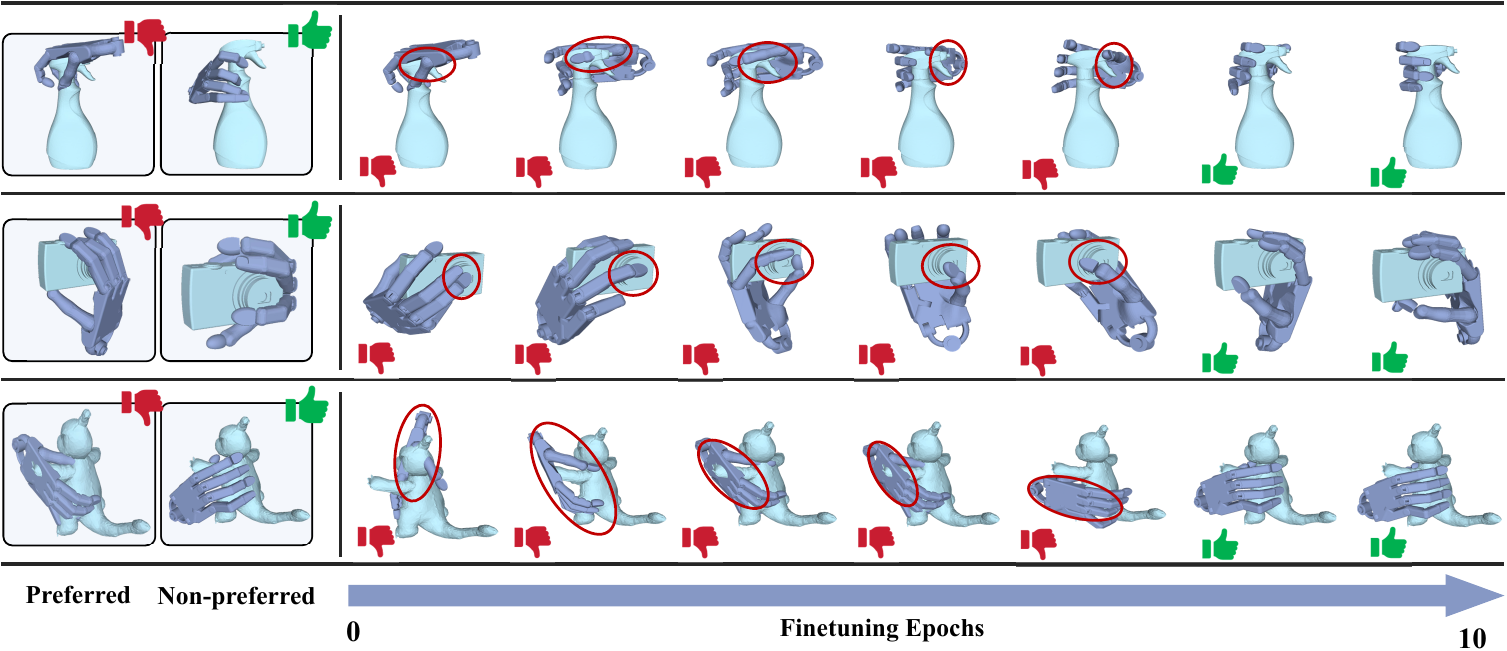}
	\centering
	\caption{Evolution of robotic grasp preferences during efficient feedback-driven finetuning across 10 epochs. Top row illustrates the adjustment from hand occlusion to clear nozzle visibility. Middle row demonstrates the transition from lens obstruction to an unobstructed camera view. The bottom row shows the evolution from a top-down grasping approach to a bottom-up one, while simultaneously mitigating physical impacts.}
    \label{Evolution_grasping}
    \vspace{-2ex}
\end{figure*}

\section{Experiments}
We first simply present experimental setup which includes datasets, evaluation metrics and baselines in Sec.~\ref{Experimental_Setup}. More details of setup including datasets and implementation details are in Appendix C. Then we provide the quantitative and qualitive results of EvolvingGrasp in Sec.~\ref{Main Results}, followed by ablation study about different modules and hyperparameter analyses in Sec.~\ref{Ablation Study}. Next, we demonstrate the evolutionary improvement of EvolvingGrasp through an experiment starting from a suboptimal model trained on a degraded dataset in Sec.~\ref{Degraded_Dataset}. Finally, We will demonstrate EvolvingGrasp in the real-world deployment in Sec. \ref{Deployment}.

\subsection{Experimental Setup}
\textbf{Datasets.} We conduct experiment on four datasets, including DexGraspNet~\cite{wang2023dexgraspnet}, Multidex~\cite{li2023gendexgrasp}, Realdex~\cite{liu2024realdex}, DexGRAB~\cite{taheri2020grab} respectively. 



\noindent \textbf{Evaluation Metrics.} We use four metrics to evaluate the grasping performance.
Success rate \textbf{Suc.6} measures the proportion of grasping poses where the object's displacement does not exceed 2 cm in all six axial directions ($\pm X$, $\pm Y$, $\pm Z$), evaluating multi-directional stability. \textbf{Suc.1} measures the proportion where displacement does not exceed 2cm in at least one direction, assessing single-direction stability. \textbf{Pen.} indicates the maximum penetration depth (mm) between the hand and the object, with lower values suggesting more physically plausible grasps. Above metrics are calculated in the IssacGym simulator~\cite{makoviychuk2021isaac} with settings consistent with SceneDiffuser~\cite{scenediffuser}.
For efficiency, \textbf{Time} refers to the computational time required to generate grasping poses for a batch of objects.

\noindent \textbf{Baselines.} 
Some generative-based methods such as UniDexGrasp \cite{xu2023unidexgrasp}, GraspTTA \cite{grasptta}, SceneDiffuser \cite{scenediffuser}, UGG \cite{lu2023ugg}, and DexGrasp Anything \cite{graspanyting} are compared on four benchmark datasets.

\subsection{Main Results}

To validate the effectiveness of our method in continuously enhancing grasping performance, we conduct quantitative evaluations on six randomly selected objects from each dataset, measuring both the Suc.6 metric and the degree of penetration. The results are illustrated in Fig.~\ref{DPO_6_performance}, indicating that as the number of inferences increases, the Suc.6 metric steadily improves. Although penetration exhibits some fluctuations due to increased contact area between the hand and the object, it still follows a downward trend.
Additionally, we present qualitative results in Fig.~\ref{Evolution_grasping}, demonstrating how our method integrates human preferences during inference. This enables the selection of more favorable poses and progressively refines grasping poses to better align with human preferences through iteration. For instance, early generated poses may obstruct the camera lens, but after several rounds of preference-based finetuning, the model learns to produce poses that avoid blocking the lens.

We conduct the experiment with comparing with other dexterous grasp generation methods in Tab.~\ref{tab:comparison}.
Compared to other generative-based approaches, our method excels in generating high-quality, physically plausible grasping poses while significantly reducing computational time. It is worth highlighting that the proposed method achieves \textbf{30x speedup} (32s$\sim$38s v.s. 0.73s$\sim$2.7s) compared to these SOTAs.
After fine-tuning, our approach achieves superior performance by leveraging its evolutionary capability through iterative refinement. Notably, our preference fine-tuning guides the model to generate poses that better align with human preferences, which may, to some extent, reduce the diversity of the generated results.
We also evaluate the trade-off between efficiency and grasp quality during preference alignment with different numbers of inference steps. While increasing the steps improves performance, it inevitably incurs additional computational costs.
Moreover, even without guidance during sampling, we can achieve comparable results in real time. 

\label{Main Results}

\begin{table*}[ht!]
\caption{Grasping performance in terms of Suc.6, Suc.1, and Pen. comparison across different methods and datasets. ``Step" refers to inference timestep.
Bold values highlight the best results and underlined values indicate the runner-up.
}
\centering
\resizebox{\linewidth}{!}{
\begin{tabular}{@{}l@{\hspace{0.2cm}}|*{16}{@{\hspace{0.1cm}}c}|c@{}}
\toprule
\multirow{2}{*}{\hspace{0.2cm} \makecell{\centering \diagbox[width=2.5cm, height=0.9cm, linewidth=0.8pt]{\textbf{Method}}{\textbf{Dataset}}}} & \multicolumn{4}{c}{DexGraspNet} &  \multicolumn{4}{c}{MultiDex} & \multicolumn{4}{c}{RealDex} & \multicolumn{4}{c|}{DexGRAB} & \multirow{2.5}{*}{\textbf{Time $\downarrow$}} \\ 
\cmidrule(lr){2-5} \cmidrule(lr){6-9} \cmidrule(lr){10-13} \cmidrule(lr){14-17}
&\textbf{Suc.6 $\uparrow$} & \textbf{Suc.1 $\uparrow$} & \textbf{Pen. $\downarrow$} & &\textbf{Suc.6 $\uparrow$} & \textbf{Suc.1 $\uparrow$} & \textbf{Pen. $\downarrow$} & & \textbf{Suc.6 $\uparrow$} & \textbf{Suc.1 $\uparrow$} & \textbf{Pen. $\downarrow$} & & \textbf{Suc.6 $\uparrow$} & \textbf{Suc.1 $\uparrow$} & \textbf{Pen. $\downarrow$} & \\ \midrule
UniDexGrasp~\cite{xu2023unidexgrasp}  & 33.9 & 70.1 & 31.9 & & 21.6 & 47.5 & 13.5 & & 27.1 & 59.4 & 39.0 & & 20.8 & 55.8 & 37.4 & & $\underline{0.46}^{\pm 0.11}$ \\
GraspTTA~\cite{grasptta}  & 18.6 & 67.8 & 24.5 & & 30.3 & 62.8 & 19.0 & & 13.3 & 46.4 & 40.1 & & 14.4 & 51.0 & 51.4 & & $10.41^{\pm 0.32}$ \\
SceneDiffuser~\cite{scenediffuser}  & 26.6 & 66.9 & 31.0 & & 69.8 & 85.6 & 14.6 & & 21.7 & 56.1 & 42.0 & & 39.1 & 85.0 & 41.1 & & $3.41^{\pm 0.13}$ \\
UGG~\cite{lu2023ugg}  & 46.9 & 79.0 & 25.2 & & 55.3 & 93.4 & \underline{10.3} & & 32.7 & 63.4 & 34.4 & & 42.7 & 90.6 & 33.2 & & $38.34^{\pm 2.31}$ \\ 

DexGrasp Any.~\cite{graspanyting}  & 53.6 & 90.4 & 21.5 & & 72.2 & 96.3 & \textbf{9.6} & & 34.6& 71.2 & 23.1 & & 56.5 & 91.8 & 28.6 & & $32.91^{\pm 1.34}$ \\\midrule

Ours w/o HPO (2-step)  & 60.8 & 91.0 & 19.2 & & 65.6 & 97.5 & 15.2 & & 41.9 & 75.4 & \textbf{19.5} & & 52.2 & 93.9 & 25.1 & & $0.73^{\pm 0.03}$ \\

Ours (2-step)  & 62.4 & 90.5 & 19.3 & & 65.9 & 97.2 & 15.3 & & 44.0 & 77.8 & \underline{19.7} & & 53.3 & 92.5 & 24.3 & & $0.73^{\pm 0.03}$ \\

Ours w/o HPO (4-step)  & 63.8 & \underline{93.0} & 17.4 & & 75.3 & 97.1 & 13.1 & & 51.6 & 82.9 & 20.5 & & 55.6 & 96.0 & 23.8 & & $1.41 ^{\pm 0.07}$ \\

Ours (4-step) & \underline{65.2} & 92.7 & 17.2 & & \underline{76.8} & \underline{98.4} & 13.0 & & 50.6 & 82.5 & 20.3 & & \underline{57.7} & 95.2 & 23.7 & & $1.41 ^{\pm 0.07}$ \\

Ours w/o HPO (8-step)  & \underline{65.2} & \textbf{93.5} & \underline{16.2} & & 75.6 & \textbf{98.7} & 12.2 & & \underline{63.6} & 86.6 & 21.9 & & 56.8 & \textbf{96.8} & \underline{23.1} & & $2.71 ^{\pm 0.08}$ \\

Ours (8-step)  & \textbf{65.4} & 92.3 & \textbf{15.9} & & \textbf{80.3} & \textbf{98.7} & 12.3 & & \textbf{64.4} & \textbf{89.1} & 21.8 & & \textbf{60.8} & \underline{96.4} & \textbf{22.3} & & $2.71 ^{\pm 0.08}$ \\
\midrule
Real-time (2-step) & 55.2 & 90.5 & 20.0 & & 63.7 & 95.0 & 13.8 & & 46.5 & 78.9 & 21.4 & & 48.9 & 93.3 & 24.8 & & $\textbf{0.06} ^{\pm 0.01}$ \\
Real-time (4-step) & 59.9 & 90.6 & 19.8 & & 64.3 & 96.5 & 11.6 & & 58.2 & \underline{87.0} & 21.1 & & 55.4 & 95.4 & 24.1 & & $0.10 ^{\pm 0.02}$ \\
\bottomrule
\end{tabular}}
\vspace{-1ex}
\label{tab:comparison}
\end{table*}

\begin{table}[htbp]

\caption{Ablation study on incorporating physical constraints during both training and sampling stages and the LLM module. The evaluation is conducted on Multidex when timestep is 4.} 
\centering
\resizebox{\linewidth}{!}{
\begin{tabular}{c|cccc|cccc} 
\toprule
 & \textbf{CM} & \textbf{PGD} & \textbf{PGS} & \textbf{HPO} & \textbf{Suc.6 $\uparrow$} & \textbf{Suc.1 $\uparrow$} & \textbf{Pen. $\downarrow$} & \textbf{Time $\downarrow$} \\ 
\midrule
a & \checkmark &   &   &   & 60.0 & 94.6 & 14.0 & $0.10 ^{\pm 0.02}$ \\

b & \checkmark & \checkmark & &  & 64.3 & 96.5  & 12.5 & $0.10 ^{\pm 0.02}$   \\
c & \checkmark & & \checkmark &  & 66.2 & 95.6  & 14.9 & $1.41 ^{\pm 0.07}$   \\
d & \checkmark & \checkmark & & \checkmark  & 67.5  & 96.5 & 11.9 & $0.10 ^{\pm 0.02}$\\
e & \checkmark & \checkmark & \checkmark &   & 75.3 & 97.1 & 13.1 & $1.41 ^{\pm 0.07}$ \\
f & \checkmark & \checkmark & \checkmark & \checkmark  & 76.8  & 98.4 & 13.0 & $1.41 ^{\pm 0.07}$ \\
\bottomrule
\end{tabular}}
\vspace{-2ex}
\label{tab:ablation_study}
\end{table}

\subsection{Ablation Study}
\subsubsection{Ablation on Different Modules} 
We investigate the impact of different modules in the physics-aware consistency model, \textbf{P}hysical constraints \textbf{G}uidance in \textbf{D}istillation and \textbf{S}ampling from which we dubbed as \textbf{PGD} and \textbf{PGS}. We also consider the performance of HPO with and without physical guidance during sampling on test split of the multidex dataset. The results presented in Tab.~\ref{tab:ablation_study} demonstrate the key role of physical constraints and preference finetuning to successful grasping pose generation during preference alignment. 

\label{Experimental_Setup}
\begin{figure}[ht]
	\centering
\includegraphics[width=0.85\linewidth]{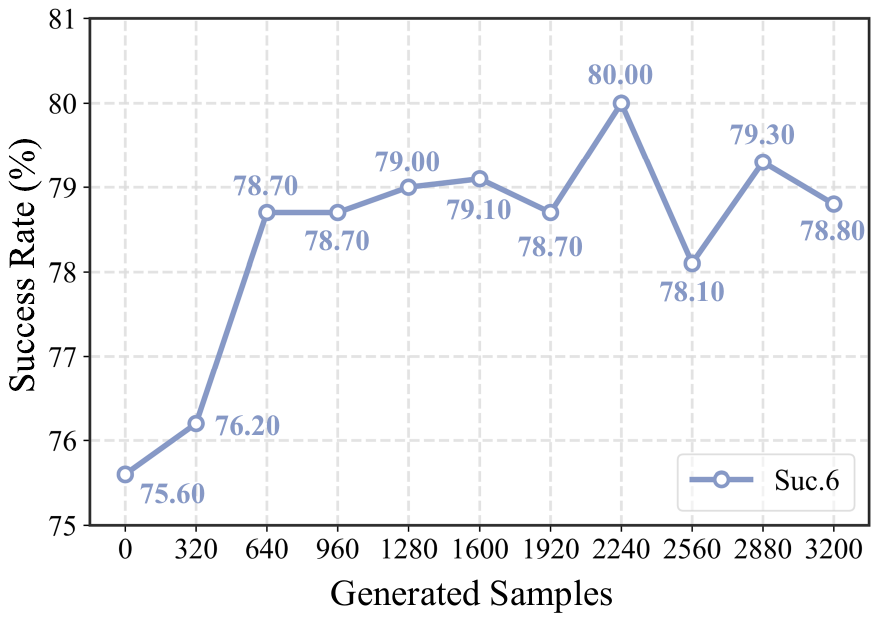}
	\centering
	\caption{The relationship between the success rate of EvolvingGrasp and the number of generated samples.}
    \label{succ_epoch}
    \vspace{-2ex}
\end{figure}

\subsubsection{Ablation on Evolutionary Finetuning}
HPO not only effectively aligns grasping poses with human preferences at the individual object level but also enhances performance across the entire dataset. 
During inference across the entire dataset, we collect both successful and failed grasping samples for each object, utilizing Eq. (\ref{loss_upperbound}) to perform lightweight finetuning of the entire model. Fig.~\ref{succ_epoch} illustrates the relationship between EvolvingGrasp's Suc.6 metric and the number of generated samples, showing that our method continuously enhances grasping performance as more samples are generated. Further ablation studies on the impact of different hyperparameters are provided in Appendix D.2.

\label{Ablation Study}

\begin{figure}[t]
	\centering
\includegraphics[width=0.85\linewidth]{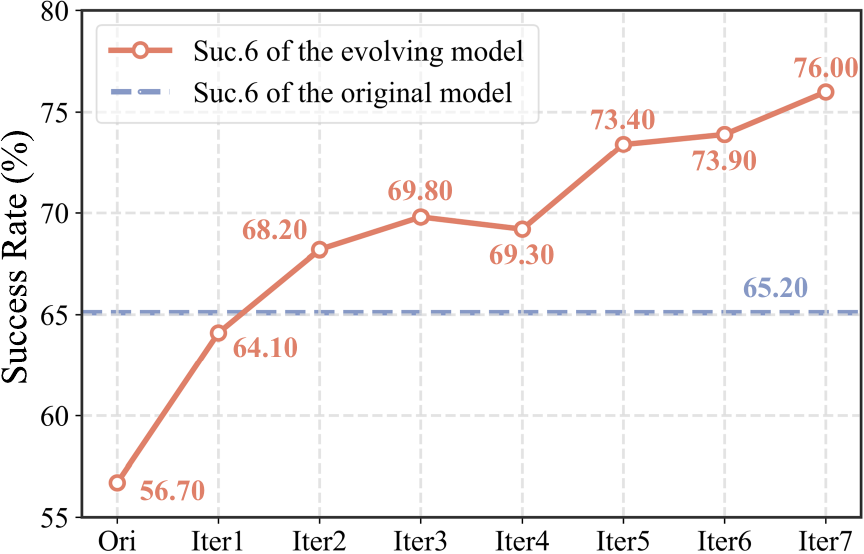}
	\centering
	\caption{The variation of the mean grasping success rate of the evolving model for randomly selected objects in the multidex dataset with increasing finetuning epochs. The blue dashed line represents the grasping success rate under the original model.}
    \label{bad_to_good}
    \vspace{-2ex}
\end{figure}

\subsection{Training from a Degraded Dataset}
In this subsection, we explore the feasibility of enhancing model performance through feedback-driven finetuning for single-object grasping tasks, particularly when trained on suboptimal data.
To begin with, we randomly add the noise on the hand pose parameter of the original multidex dataset. This perturbation creates a degraded dataset, which we then use to train and distill a suboptimal model. During the inference phase of this suboptimal model, we collect successfully grasping poses as positive samples and all other poses as negative samples. We then employ Eq.~\ref{loss_upperbound} to finetune the whole model with LoRA~\cite{hu2022lora}. Fig.~\ref{bad_to_good} illustrates the change in the average grasping success rate of the suboptimal model on randomly sampled objects from the multidex dataset as the number of fine-tuning epochs increases. 
The results indicate that we continuously improve the success rate during the evolutionary adaptation process and ultimately outperform the accuracy of the original model.
\label{Degraded_Dataset}
\begin{figure}[t]
	\centering  \includegraphics[width=1\linewidth]{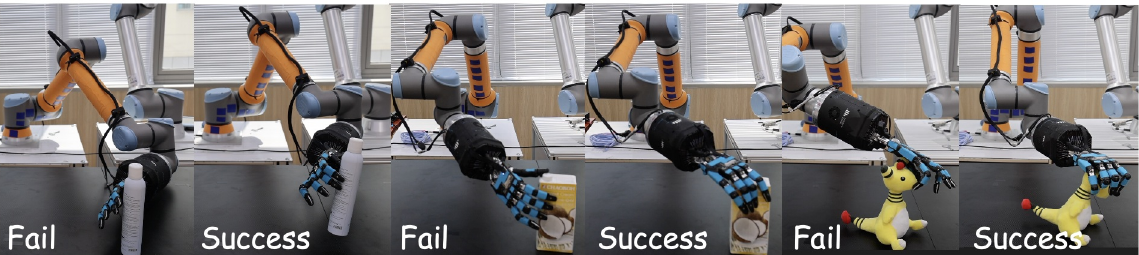}
	\centering
	\caption{Real-world deployment on Shadow Hand.}
    \label{real}
    \vspace{-3ex}
\end{figure}
\subsection{Real World Deployment}
To verify that EvolvingGrasp can continuously enhance grasping performance and align with human preferences in real world, we deploy our model on a real ShadowHand robot, as shown in Fig.~\ref{real}. The pre-grasping motion trajectory is generated based on RealDex~\cite{liu2024realdex}. Real-world experiments demonstrate that our method achieves successful grasping through several efficient preference finetuning when the initial grasp attempt fails. Additional video demonstrations can be found in supplementary materials.

\label{Deployment}

%% file: sec/5_conclusion.tex
\section{Conclusion}
We propose EvolvingGrasp, an evolutionary grasp generation method through efficient preference alignment. 
HPO is introduced to allow the model to continuously align the performance with preference signals. 
We design the Physics-Aware Consistency Model to achieve both rapid inference and efficient preference finetuning while maintaining physical plausibility.
Extensive experiments across four benchmarks demonstrate that our method achieves state-of-the-art results. Furthermore, we deploy our model on a real ShadowHand robot to validate its evolutionary capability in real-world scenarios.

\label{sec:conclusions}

%% file: sec/X_suppl.tex
\clearpage

\onecolumn
\appendix
\noindent \textbf{\Large Appendix}
\section{Pseudo Code of EvolvingGrasp}
Pseudo Code of EvolvingGrasp is shown in Algorithm 1 and 2.

\renewcommand{\thealgorithm}{1}
    \begin{algorithm*}[!htbp]
        \caption{Physics-Aware Sampling and Handpose-Wise Preference Optimization} 
        \label{online}
        \begin{algorithmic}[1]
            \Require Number of inference timesteps $T$, number of finetuning epochs $E_{ft}$, number of objects $K$, physical-aware consistency model ${\epsilon}_\theta$, test batchsize $B$, test time sequences $S \in \left\{\tau_{i} \mid \tau_{0}=0, \tau_{N-1}=T, \tau_{i}<\tau_{i+1} \text { for } i=0,1, \ldots, N-1\right\}$, differentiable functions $c_{skip}$ and $c_{out}$, gradient guidance weight $\left \{  \gamma_{i} \right \}_{i=1}^{m}$.
            \State Copy the parameters of consistency model $ {\epsilon}_\mathrm{ref} =  {\epsilon}_\theta$ and set $ {\epsilon}_{\mathrm{ref}}$ to have \texttt{requires\_grad = False}.
            \For {$e=1:E_{ft}$}
                \State \texttt{\# Sample grasping poses}
                \For{$k=1:K$}
                    \State Choose an object ${O_k}$ and sample ${x}_{T}\sim \mathcal{N}({0}, \textbf{I})$
                    \For{$i=1:B$}
                        \For{$n=N-1:0$}
                    \State $F_{\theta}({x}^i_{k,\tau_{n}}, \tau_{n}, O_{k})=\frac{1}{\sqrt{\bar{\alpha}_{\tau_{n}}}} \left({x}^i_{k,\tau_{n}}-\sqrt{1-\bar{\alpha}_{\tau_{n}}} \epsilon_{\theta}\left({x}^i_{k,\tau_{n}, }, \tau_{n}, O_k\right)\right)$
                    \State $f_{\theta}({x}^i_{k,\tau_{n}}, \tau_{n}, O_{k})=c_{\text {skip }}(\tau_{n}) {x}^i_{k,\tau_{n}}+c_{\text {out }}(\tau_{n}) F_{\theta}({x}^i_{k,\tau_{n}}, \tau_{n}, O_k)$
                            \State \texttt{\# Sampling with \textcolor{red}{Gradient Guidance}}:
                            
                            \State $\hat{\mu}_{\theta}( {x}^i_{k,\tau_{n}},\tau_{n}, {O_k})= \sqrt{\bar{\alpha}_{\tau_{n-1}}}f_{\theta}({x}^i_{k,\tau_{n}}, \tau_{n}, O_{k}) 
                            + \textcolor{red}{\sum_{i=1}^{m} \gamma_{i} \nabla_{x_{\tau_{n}}} L_{PA_{i}}\left( F_{\theta}({x}^i_{k,\tau_{n}}, \tau_{n}), \epsilon_{\theta} \right)}$ 
                            \State $\sigma_{\tau_{n}}=\sqrt{1-\bar{\alpha}_{\tau_{n-1}}}$
                            \State $ {x}^i_{k,\tau_{n-1}}= \hat{\mu}_{\theta}( {x}^i_{k,\tau_{n}},\tau_{n}, {O_k}) +\sigma_{\tau_{n}}  {\epsilon}, \;\;\;\;  {\epsilon}\sim \mathcal{N}({0}, \textbf{I})$
                    \EndFor
            \EndFor
                \State \texttt{\# Select the Preferred Grasp Poses}
                    \For{$i=0:B$}
                        \If{$x_{0}^{i}$ grasp object $O_{k}$ matches human preference}
                        \State $h_i=1$
                        \Else
                        \State $h_i=-1$
                        \EndIf
                    \EndFor
                \State \texttt{\# Efficiently Feedback-driven Finetuning}
            \For{$n=N-1:0$}
                \State {\texttt{\# Utilizing \textcolor{red}{Fewer Timesteps} for Preference Alignment.}}
                \For{$i=1:B$}
                \State \texttt{with grad}: 
                \State ${\mu}_{\theta}( {x}^i_{k,\tau_{n}},\tau_{n}, {O_k})= \sqrt{\bar{\alpha}_{\tau_{n-1}}}f_{\theta}({x}^i_{k,\tau_{n}}, \tau_{n}, O_{k})$, ${\mu}_{\mathrm{ref}}( {x}^i_{k,\tau_{n}},\tau_{n}, {O_k})= \sqrt{\bar{\alpha}_{\tau_{n-1}}}f_{\mathrm{ref}}({x}^i_{k,\tau_{n}}, \tau_{n}, O_{k})$
                \State $\pi_{\theta}\left({x}^i_{k,\tau_{n-1}} \mid {x}^i_{k,\tau_{n}}, O_k\right) = 
                \frac{1}{\sqrt{2 \pi} \sigma_{\tau_{n}}} \text{exp} (-\frac{({x}^i_{k,\tau_{n}}-\mu_{\theta }({x}^i_{k,\tau_{n}},\tau_{n}, O_k))^{2}}{2\sigma_{\tau_{n}}^{2}} )$ 
                \State $\pi_\mathrm{ref}\left({x}^i_{k,\tau_{n-1}} \mid {x}^i_{k,\tau_{n}}, O_k\right) = 
                \frac{1}{\sqrt{2 \pi} \sigma_{\tau_{n}}} \text{exp} (-\frac{({x}^i_{k,\tau_{n}}-\mu_{\mathrm{ref}}({x}^i_{k,\tau_{n}},\tau_{n}, O_k))^{2}}{2\sigma_{\tau_{n}}^{2}} )$
                \EndFor
                \State Update $\theta$ using gradient descent with \textcolor{red}{LoRA}: 
                \[
                \nabla_\theta \log \sigma(\sum_{i=1}^{B} h_i \beta  \log \frac{\pi_\theta( {x}^i_{k,\tau_{n-1}}| {x}^i_{k,\tau_{n}},\tau_{n}, {O_k})}{\pi_\mathrm{ref}( {x}_{k,\tau_{n-1}}| {x}^i_{k,\tau_{n}},\tau_{n}, {O_k})}) 
                \]
                \EndFor
            \EndFor
            \EndFor  
        \end{algorithmic}
    
    \end{algorithm*}

\renewcommand{\thealgorithm}{2}
    \begin{algorithm*}[!htbp]
        \caption{Physical-Aware Distillation} 
        \label{online}
        \begin{algorithmic}[1]
            \Require Training dataset $D_{t}$, number of training epochs $E_{t}$, learning rate $\eta$, pre-trained diffusion model $ {\epsilon}_\theta$, number of timesteps $T_{dm}$, distance metric $d(\cdot,\cdot)$, EMA rate $\mu$, noise schedule $\left \{ \alpha_{t} \right \}_{t=1}^{T_{dm}}$, physics-aware constraints weights $\left \{  \alpha_{i} \right \}_{i=1}^{m}$.
            \State Copy the parameters of the pre-trained diffusion model as the target network $ {\epsilon}_\mathrm{\theta^{\prime}} =  {\epsilon}_\theta$
            \For {$e=1:E$}
                \For{$k=1:K$}
                    \State Choose an object ${O_k}$ and sample ${x}_{0} \sim D_{t}$, $n \sim \mathcal{U}[1, N]$
                    \State Sample $x_{\tau_{n}} \sim \mathcal{N}(\sqrt{\bar{\alpha}_{\tau_{n}}}x_{0}, (1-\bar{\alpha}_{\tau_{n}})\textbf{I})$
                    \State $F_{\theta}(x_{\tau_{n}}, \tau_{n}, O_{k})=
                    \frac{1}{\sqrt{\bar{\alpha}_{\tau_{n}}}} \left(x_{\tau_{n}}-\sqrt{1-\bar{\alpha}_{\tau_{n}}} \epsilon_{\theta}\left(x_{\tau_{n}}, \tau_{n},O_{k}\right)\right)$
                    \State $\hat{x}_{\tau_{n-1}}^{\ast} = \sqrt{\bar{\alpha}_{\tau_{n-1}}}F_{\theta}(x_{\tau_{n}}, \tau_{n}, O_{k}) + \sqrt{1-\bar{\alpha}_{\tau_{n-1}}} \epsilon, \;\;\;\;  {\epsilon}\sim \mathcal{N}({0}, \textbf{I})
                    $ 
                    \State $\mathcal{L}_{PAD}=\mathbb{E}\left[d\left(f_{\theta}\left(x_{\tau_{n}}, \tau_{n}\right), f_{\theta^{\prime}}\left(\hat{x}_{\tau_{n-1}}^{\ast }, \tau_{n-1}\right)\right)\right]+
                    \sum_{i=1}^{m} \alpha_{i} L_{PA_{i}}\left(F_{\theta}(x_{\tau_{n}}, \tau_{n}), \epsilon_{\theta}\right)$
                    \State ${\theta} \leftarrow {\theta}-\eta \nabla_{{\theta}} \mathcal{L}_{PAD}$
                    \State ${\theta}^{\prime} \leftarrow \operatorname{stopgrad}\left(\mu {\theta}^{\prime}+(1-\mu) {\theta}\right)$
                    
            \EndFor
            \EndFor  
        \end{algorithmic}
    
    \end{algorithm*}

\section{Proof}
Defined on the new path, the proof of the upper bound is as follows:
\begin{equation}
\begin{aligned}
L_{\mathrm{BT}}(\theta) 
=& -\mathbb{E}_{x_0^{w,l} \sim \mathcal{D}} \log \sigma \Bigg( 
\quad  \mathop{\beta \mathbb{E}}_{\substack{
    x_{1:T}^{w,l} \sim \pi_\theta(x_{1:T}^{w,l} \mid x_0^{w,l})
}} \left[ 
    \log \frac{\pi_\theta(x_{0:T}^w)}{\pi_{\mathrm{ref}}(x_{0:T}^w)} 
    - \log \frac{\pi_\theta(x_{0:T}^l)}{\pi_{\mathrm{ref}}(x_{0:T}^l)} 
\right] \Bigg) \\
= & -\mathbb{E}_{x_0^{w,l} \sim \mathcal{D}} \log \sigma \Bigg( 
\quad  \mathop{\beta \mathbb{E}}_{\substack{
    x_{1:T}^{w,l} \sim \pi_\theta(x_{1:T}^{w,l} \mid x_0^{w,l})
}} \left[ 
    \sum_{n=1}^{N} \log \frac{\pi_\theta\left({x}_{\tau_{n-1}}^{w} \mid {x}_{\tau_{n}}^{w}\right)}{\pi_{\mathrm{ref}}\left({x}_{\tau_{n-1}}^{w} \mid {x}_{\tau_{n}}^{w}\right)} 
    - \sum_{n=1}^{N} \log \frac{\pi_\theta \left({x}_{\tau_{n-1}}^{l} \mid {x}_{\tau_{n}}^{l}\right)}{\pi_{\mathrm{ref}}\left({x}_{\tau_{n-1}}^{l} \mid {x}_{\tau_{n}}^{l}\right)} 
\right] \Bigg) \\
= & -\mathbb{E}_{x_0^{w,l} \sim \mathcal{D}} \log \sigma \Bigg( 
\quad  \mathop{\beta \mathbb{E}}_{\substack{
    x_{1:T}^{w,l} \sim \pi_\theta(x_{1:T}^{w,l} \mid x_0^{w,l})
}} N\mathbb{E}_{n} \left[ 
\log \frac{\pi_\theta\left({x}_{\tau_{n-1}}^{w} \mid {x}_{\tau_{n}}^{w}\right)}{\pi_{\mathrm{ref}}\left({x}_{\tau_{n-1}}^{w} \mid {x}_{\tau_{n}}^{w}\right)} 
- \log \frac{\pi_\theta \left({x}_{\tau_{n-1}}^{l} \mid {x}_{\tau_{n}}^{l}\right)}{\pi_{\mathrm{ref}}\left({x}_{\tau_{n-1}}^{l} \mid {x}_{\tau_{n}}^{l}\right)} 
\right] \Bigg) \\
= & -\mathbb{E}_{x_0^{w,l} \sim \mathcal{D}} \log \sigma \Bigg( 
\quad \mathop{N \beta \mathbb{E}}_{\substack{n, x_{\tau_{n}}^{w,l} \sim \pi_{\theta}\left(x_{\tau_{n}}^{w,l} \mid x_{0}^{w,l}\right)}} \left[ 
\log \frac{\pi_\theta\left({x}_{\tau_{n-1}}^{w} \mid {x}_{\tau_{n}}^{w}\right)}{\pi_{\mathrm{ref}}\left({x}_{\tau_{n-1}}^{w} \mid {x}_{\tau_{n}}^{w}\right)} 
- \log \frac{\pi_\theta \left({x}_{\tau_{n-1}}^{l} \mid {x}_{\tau_{n}}^{l}\right)}{\pi_{\mathrm{ref}}\left({x}_{\tau_{n-1}}^{l} \mid {x}_{\tau_{n}}^{l}\right)} 
\right] \Bigg) \\
\leq &-  \mathop{\mathbb{E}}_{\substack{{x}_{0}^{w,l} \sim \mathcal{D}, t \sim \mathcal{U}(0, T), \\
n, x_{\tau_{n}}^{w} \sim \pi_{\theta}\left(x_{\tau_{n}}^{w} \mid x_{0}^{w}\right)}}
\log \sigma\left(\beta \log \frac{\pi_{\theta}\left({x}_{\tau_{n-1}}^{w} \mid {x}_{\tau_{n}}^{w}\right)}{\pi_{\mathrm{ref}}\left({x}_{\tau_{n-1}}^{w} \mid {x}_{\tau_{n}}^{w}\right)}-\beta \log \frac{\pi_{\theta}\left({x}_{\tau_{n-1}}^{l} \mid {x}_{\tau_{n}}^{l}\right)}{\pi_{\mathrm{ref}}\left({x}_{\tau_{n-1}}^{l} \mid {x}_{\tau_{n}}^{l}\right)}\right)
\end{aligned}
\label{proof_upper_bound}
\end{equation}
where the last inequality is based on Jensen’s inequality and $-\log \sigma(\cdot)$ is a strict convex function. Therefore, we use the new objective \ref{proof_upper_bound} to optimize the whole model with LoRA~\cite{hu2022lora}.

\label{proof}

\begin{table*}[h!]
\caption{Cross-dataset evaluation results. The highest performances are highlighted in \textbf{bold}, while the second-highest performances are indicated with \underline{underline}.
}
\centering
\resizebox{\linewidth}{!}{
\begin{tabular}{@{}l@{\hspace{0.2cm}}*{14}{@{\hspace{0.1cm}}c}@{}}
\toprule
\textbf{Testing Dataset} & \multicolumn{3}{c}{DexGraspNet}  & \multicolumn{3}{c}{MultiDex} & \multicolumn{3}{c}{RealDex} & \multicolumn{3}{c}{DexGRAB}\\ 
\cmidrule(lr){2-4} \cmidrule(lr){5-7}\cmidrule(lr){8-10}\cmidrule(lr){11-13}
\textbf{Training Dataset} &\textbf{Suc.6 $\uparrow$} & \textbf{Suc.1 $\uparrow$} & \textbf{Pen. $\downarrow$} & \textbf{Suc.6 $\uparrow$} & \textbf{Suc.1 $\uparrow$} & \textbf{Pen. $\downarrow$} & \textbf{Suc.6 $\uparrow$} & \textbf{Suc.1 $\uparrow$} & \textbf{Pen. $\downarrow$} & \textbf{Suc.6 $\uparrow$} & \textbf{Suc.1 $\uparrow$} & \textbf{Pen. $\downarrow$} &  \\ 
\midrule
DexGraspNet &\underline{65.2} &\underline{92.7}& 17.2  &73.4 &97.1 &\textbf{9.7} &\textbf{54.1}& \textbf{90.1}& \textbf{19.4} &\underline{58.1} &94.3 &20.6 \\ 
MultiDex  &\textbf{67.6} &\textbf{94.0} &\textbf{19.5} &\textbf{76.8} &\underline{98.4} &13.0 &51.9 &\underline{88.6} &\textbf{19.4} &\textbf{65.6} &\textbf{96.8} &\underline{19.5}  \\ 
RealDex  &52.2 &81.1 &20.7 &51.5 &88.1 &14.0 & 50.6 &82.5 &20.3 &46.0 &80.0 &\textbf{18.3}    \\ 
DexGRAB  &64.9 &92.6 &\underline{17.1} &\underline{75.3} &99.3 &\underline{9.9} &\underline{53.1} &88.5 &\underline{19.7} &57.7 &\underline{95.2} &23.7  \\ 
\bottomrule
\end{tabular}}
\label{cross_dataset}
\end{table*}

\twocolumn

\section{Details of Experimental Setup}
\label{Details_of_Experimental_Setup}

\subsection{Dataset Setups}
\label{sup-dataset}
DexGraspNet is a large-scale dataset for dexterous grasping, comprising 1.32 million grasp samples across 5,355 objects from 133 diverse categories. While its optimization-based generation ensures high quality and diversity, its applicability in real-world scenarios is limited.

In contrast, MultiDex focuses on a smaller set of 58 everyday objects but offers a rich variety of grasping poses for each object. This makes it an ideal dataset for studying the diversity of grasping configurations and developing methods that can generate a wide range of effective grasps for common objects.

Realdex shifts the focus to real-world applications by capturing natural human grasping behaviors. It contains 59,000 samples across 52 objects, making it highly suitable for training robots to learn human-like grasping poses. Although it covers fewer object categories, its real-world grounding allows it to effectively validate the generalization and practicality of dexterous grasping methods in real environments.

DexGRAB, derived from human hand interaction data, provides over 1.64 million grasp samples across 51 distinct objects. It offers rich grasping patterns and natural interaction behaviors, making it a valuable resource for understanding human grasping strategies. Similar to DexGraspNet, DexGRAB's data quality is high after filtering, but its real-world applicability may also face some limitations due to its primarily simulation-based nature.

Together, these datasets offer a range of strengths and limitations, from the large-scale optimization-based approaches of DexGraspNet and DexGRAB to the real-world grounding of Realdex and the diversity-focused MultiDex. Each dataset contributes unique insights and challenges to the field of dexterous grasping research.

\subsection{Implementation Details}
Our EvolvingGrasp contains distillation and sampling, which are implemented using Pytorch~\cite{paszke2019pytorch} platform in one NVIDIA Tesla A40 GPU. 
In the distillation process, we train EvolvingGrasp for 1,000 epochs with a batch size of 1,200. During both the distillation and preference finetuning processes, the initial learning rate is set to 0.00001. For the distillation process, the learning rate remains unchanged. During inference, the success rate of the generated grasping poses is firstly evaluated. If the success rate improves, the learning rate is adaptively reduced, otherwise, it is increased accordingly. Additionally, the adjustment of the learning rate is constrained within a predefined threshold range to ensure it remains within reasonable bounds. The sampling and preference optimization processes are implemented in test split of each corresponding dataset.
\label{Implementation_Details}

\section{Additional Experiments}

\subsection{Performance of Cross Dataset}
We conducted cross-validation experiments on four datasets with our method and one dataset with four methods. The results with four datasets are shown in Table~\ref{cross_dataset}, which demonstrate that the Physics-Aware Consistency Model trained on the Multidex dataset achieved the best performance when tested on the other datasets. The model trained and tested on the DexGRAB and DexGraspNet datasets showed moderate performance. Since Realdex is a real-world dataset with relatively lower quality, the performance of the model trained and tested on Realdex was relatively worse. The results with four methods are shown in Table \ref{cross_dataset_eva}, which illustrate that our methods can significantly improve the grasping performance on the realdex dataset compared with other methods.

\begin{table}
\caption{Evaluating Cross-Dataset Generalization. Model performance is compared on RealDex, with training on DexGraspNet.}
\centering
\begin{tabular}{@{}l|@{\hspace{0.1cm}}*{3}{@{\hspace{0.1cm}}c}@{}}  
\toprule
\multirow{1}{*}{\textbf{Method}} & \textbf{Suc.6 $\uparrow$} & \textbf{Suc.1 $\uparrow$} & \textbf{Pen. $\downarrow$} \\ 
\midrule
\multirow{1}{*}{SceneDiffuser}  & 16.1 & 52.1 & \textbf{29.2} \\
\multirow{1}{*}{GraspTTA}  & 25.5 & 64.8 & \textbf{31.6}  \\
\multirow{1}{*}{UGG}  & 33.6 & 74.5 & 33.0  \\
\multirow{1}{*}{DexGrasp Any.} & 38.4 & 77.5 & \textbf{19.2}  \\
\multirow{1}{*}{Ours w/o HPO} & 52.6& 88.8 & 19.5  \\
\multirow{1}{*}{Ours} & 54.1& 90.1 & 19.4  \\
\bottomrule
\end{tabular}
\label{cross_dataset_eva}

\end{table}

\subsection{More Ablation Studies}
\label{More_Ablation_Studies}

\begin{table}[H]
\centering
\caption{Ablation study on different hyperparameters (i.e., the regularization weight $\beta$, the number of iterations per finetuning epoch $N_{ft}$). We report the results under 2, 4, and 8 steps during sampling.}
\resizebox{\linewidth}{!}{
\scriptsize
\setlength\tabcolsep{3pt}
\begin{tabular}{c|c|ccc|c|ccc}
\toprule
$T$ & $\beta$ & \textbf{Suc.6 $\uparrow$} & \textbf{Suc.1 $\uparrow$} & \textbf{Pen. $\downarrow$} & $N_{ft}$ & \textbf{Suc.6 $\uparrow$} & \textbf{Suc.1 $\uparrow$} & \textbf{Pen. $\downarrow$} \\
\midrule
\multirow{4}{*}{2} & 0.1 & 65.6 & 97.5 & 15.2 & 1 & 65.9 & 97.2 & 15.3 \\
                   & 0.5 & 63.4 & 96.8 & 15.2 & 3 & 63.4 & 97.1 & 15.1 \\
                   & 1.0 & 65.9 & 97.2 & 15.3 & 5 & 66.2 & 96.9 & 15.2 \\
                   & 2.0 & 65.9 & 97.2 & 15.3 & 10 &65.3 &97.5 &15.3 \\

\midrule
\multirow{4}{*}{4} & 0.1 & 75.9 & 97.1 & 13.1 & 1 & 76.8 & 98.4 & 13.0 \\
                   & 0.5 & 77.1 & 97.2 & 13.1 & 3 & 75.9 & 97.8 & 13.0 \\
                   & 1.0 & 76.8 & 98.4 & 13.0 & 5 & 77.5 & 97.5 & 13.2 \\
                   & 2.0 & 77.2 & 97.2 & 13.2 & 10 &75.9 &97.5 &13.1 \\

\midrule
\multirow{4}{*}{8} & 0.1 & 79.4 & 97.8 & 12.2 & 1 & 80.3 & 98.7 & 12.3 \\
                   & 0.5 & 76.8 & 97.8 & 12.1 & 3 & 76.5 & 98.4 & 12.2 \\
                   & 1.0 & 80.3 & 98.7 & 12.3 & 5 & 78.8 & 98.1 & 12.2 \\
                   & 2.0 & 78.7 & 97.5 & 12.2 & 10 &80.0 &98.1 &12.2 \\

\bottomrule
\end{tabular}
}
\label{table:ablation_diff_parameter}
\end{table}

\begin{figure}[t]
	\centering
        \includegraphics[width=0.85\linewidth]{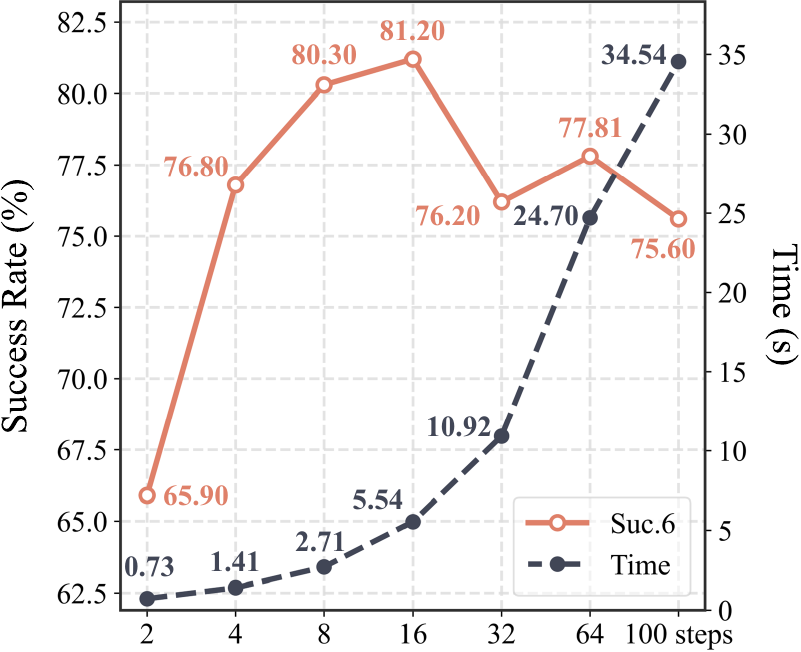}
	\centering
	\caption{The effect of different sampling steps on grasping performance. The red solid line represents the grasping success rate, while the black dashed line denotes the time consumption.
}
    \label{diff_step_abl}
\end{figure}

\noindent \textbf{Impact of different hyperparameters.} A comprehensive analysis of different hyperparameters (i.e, regularization weight $\beta$, number of finetuning $N_{ft}$ every epoch, number of timesteps $T$) to the performance during preference alignment is reported in Table \ref{table:ablation_diff_parameter} and Fig. \ref{diff_step_abl}. Table \ref{table:ablation_diff_parameter} demonstrates that when the sampling time steps is relatively small, such as 2 or 4 steps, increasing the number of finetuning iterations $N_{ft}$ and raising the value of the regularization coefficient $\beta$ can enhance the model's performance. Conversely, when the sampling steps is larger, employing fewer $N_{ft}$ and a smaller $\beta$ value helps maintain the model at a high performance level. Fig. \ref{diff_step_abl} shows that as the number of sampling steps increases, the grasping performance first improves and then declines. The highest grasping success rate is achieved when the sampling step is set to 16. The potential reason is that during the multi-step sampling process, each step introduces minor errors in noise handling. These errors may be masked in early steps but accumulate over time, eventually degrading the sample quality.

